\documentclass[sigconf]{acmart}
\usepackage{stfloats}
\usepackage{diagbox}
\usepackage{multicol}
\usepackage{nameref}
\usepackage{makecell}
\newcommand\blfootnote[1]{%
  \begingroup
  \renewcommand\thefootnote{}\footnote{#1}%
  \addtocounter{footnote}{-1}%
  \endgroup
}

\AtBeginDocument{%
  \providecommand\BibTeX{{%
    \normalfont B\kern-0.5em{\scshape i\kern-0.25em b}\kern-0.8em\TeX}}}

\setcopyright{acmcopyright}
\copyrightyear{2018}
\acmYear{2018}
\acmDOI{10.1145/1122445.1122456}

\acmConference[Woodstock '18]{Woodstock '18: ACM Symposium on Neural
  Gaze Detection}{June 03--05, 2018}{Woodstock, NY}
\acmBooktitle{Woodstock '18: ACM Symposium on Neural Gaze Detection,
  June 03--05, 2018, Woodstock, NY}
\acmPrice{15.00}
\acmISBN{978-1-4503-XXXX-X/18/06}

\begin{document}
\title{Modeling Protein Using Large-scale Pretrain Language Model}

\newcommand\xiao[1]{\textcolor{magenta}{[Yijia Xiao:  #1 ]}}
\newcommand\qiu[1]{\textcolor{magenta}{[Jiezhong Qiu:  #1 ]}}
\newcommand\li[1]{\textcolor{magenta}{[Ziang Li:  #1 ]}}
\newcommand\hsieh[1]{\textcolor{magenta}{[Chang-Yu Hsieh:  #1 ]}}
\newcommand\tang[1]{\textcolor{magenta}{[Jie Tang:  #1 ]}}

\iffalse
\fi

\setcopyright{acmcopyright}
\copyrightyear{2021}
\acmYear{2021}

\sloppy

\settopmatter{printacmref=false}
\setcopyright{none}
\renewcommand\footnotetextcopyrightpermission[1]{}
\pagestyle{plain}

\fancyhead{}

\author[ Y. Xiao, J. Qiu, Z Li, C Hsieh, and J. Tang]{
    \large Yijia Xiao$^{1~2}$, Jiezhong Qiu$^{1~2}$, Ziang Li$^{1}$, Chang-Yu Hsieh$^{3}$, Jie Tang$^{1~2}$
}

\affiliation{
    $^1$ Department of Computer Science and Technology, Tsinghua University
    \country{}
}
\affiliation{
    $^2$ Beijing Academy of Artificial Intelligence
    \country{}
}
\affiliation{
    $^3$ Tencent Quantum Lab
    \country{}
}

\email{
  {xiaoyiji18, qiujz16, li-za19}@mails.tsinghua.edu.cn
}
\email{
    kimhsieh@tencent.com
}

\email{
    jietang@tsinghua.edu.cn
}

\begin{abstract}
Protein is linked to almost every life process. Therefore, analyzing the biological structure and property of protein sequences is critical to the exploration of life, as well as disease detection and drug discovery. Traditional protein analysis methods tend to be labor-intensive and time-consuming. The emergence of deep learning models makes modeling data patterns in large quantities of data possible. Interdisciplinary researchers have begun to leverage deep learning methods to model large biological datasets, e.g. using long short-term memory and convolutional neural network for protein sequence classification. After millions of years of evolution, evolutionary information is encoded in protein sequences. Inspired by the similarity between natural language and protein sequences, we use large-scale language models to model evolutionary-scale protein sequences, encoding protein biology information in representation. Significant improvements are observed in both token-level and sequence-level tasks, demonstrating that our large-scale model can accurately capture evolution information from pretraining on evolutionary-scale individual sequences. Our code and model are available at \url{https://github.com/THUDM/ProteinLM}.
\blfootnote{$^*$Corresponding author: Jie Tang.}

\end{abstract}




\maketitle

\section{Introduction}

As an indispensable part of life activities, protein is responsible for catalysis (such as enzymes), transportation (such as hemoglobin), etc. Therefore, understanding the structure and functionality of protein is critical to the study of life science, as well as disease detection and drug discovery. Traditional protein analysis paradigms can be divided into experimental and analytical. Experimental methods usually require purification, crystallization, and X-ray crystallography. Analytical methods, like sequence alignment\cite{DBLP:journals/corr/Ma15c}, and molecular dynamics simulation \cite{GENG20191162}, tend to be incapable of handling large-scale protein data. Sequence alignment and similarity analysis leverage the idea that "structure determines properties", that sequential molecules with similar sequence order tend to have common ancestors and are relatively similar in structure and functionality. So similarity analysis often requires a large-scale annotated database, the properties of the sequence to be analyzed can be inferred from the labels of aligned sequences in the database. However, labeling such large databases requires lots of manpower and material resources. Molecular dynamics simulation (MD) and Monte Carlo (MC) simulations can be applied to protein analysis\cite{Gsponer6719}\cite{Karplus6679}, and can be quite accurate (simulation at atom-scale). However, requires a lot of computing resources and is time-consuming.

Generally speaking, most of the proteins that exist stably in nature have undergone millions of years of natural selection and evolution, and are in a low-energy stable state. The polarity of some amino acids makes certain amino acid arrangements in a lower energy state, and motifs in proteins are also made up of specific amino acid stretches folded. Such patterns can be captured by deep learning models. Researchers have explored various strategies. Inspired by Word2Vec\cite{DBLP:journals/corr/abs-1301-3781}, BioVec\cite{asgari2015continuous} proposed ProtVec for proteins GeneVec for gene sequences modeling. However, the vocabulary size grows exponentially with dependence range (n-gram), making the cost of modeling long-range dependencies unbearable (n-grams representation). With the rise of representation learning, sequence representation learning\cite{Alley589333} and transfer learning \cite{Heinzinger614313} are also introduced to protein analysis. Recent years, the emergence of the attention mechanism\cite{vaswani2017attention}, which can compute hidden representations in parallel, allows researchers to better model long sequential data. ProtTrans\cite{prottrans} also show that large-scale auto-regressive language models can model protein sequences quite well. Besides, the information encoded in an individual sequence is limited, MSA Transformer\cite{Rao2021.02.12.430858}, ESM\cite{esm} leverage sequence alignment information to model protein even better. Other research like Neural Potts Model\cite{Sercu2021.04.08.439084} obtained inspiration from the Potts model.

Thanks to the advancement of high-throughput sequencing technology, we have larger amino acid sequence databases than ever before. However, most of these data are unlabeled primary structures of proteins, the labeled sequences (like structure, stability) are relatively scarce. The amazing achievements of BERT\cite{bert} reveal the fact that data patterns can be extracted using unsupervised learning from massive unlabeled data, which inspired us to train language models on massive protein sequences. We have trained multiple large-scale models on the PFAM\cite{pfam2019} dataset, the largest with 3 billion parameters, outperforming TAPE's \cite{rao2019evaluating} performance.

\section{Related Works}
\subsection{Standardized Datasets and Tasks}
There are plenty of data in the computational proteomics field, however, current literature is fragmented in terms of unified datasets and evaluation metrics. The methods and models introduced by researchers are often evaluated on different datasets with different evaluation metrics. To solve this dilemma, TAPE\cite{rao2019evaluating} put forward a set of five biologically related tasks, including secondary structure prediction (\cite{ss}, \cite{ss_cc1}, \cite{ss_cc2}), contact prediction (\cite{rh_cc}, \cite{ss_cc1}, \cite{ss_cc2}), remote homology detection (\cite{rh_cc}), fluorescence (\cite{fluorescence}), and stability (\cite{stability}). Besides, commonly used models, like LSTM\cite{LSTM}, Transformer\cite{vaswani2017attention}, and ResNet\cite{resnet} are implemented for these tasks, serving as benchmarks for semi-supervised representation learning. One of their conclusions is that self-supervised training is beneficial for almost all models on all tasks, doubling performance in some downstream tasks. Our work is based on standardized datasets and evaluation metrics provided in TAPE\cite{rao2019evaluating}.

\subsection{Large-scale Pretraining}
The success of pretraining makes researchers wonder whether the in language model scale can always bring about improved performance. ProtTrans\cite{prottrans} is one of the representatives, the researchers trained a series of language models with tens of billions of parameters, the largest one ProtT5-XXL with 11B parameters, and achieved excellent performance on downstream tasks such as secondary structure prediction and solubility prediction.

\subsection{Efficient Pretraining of Language Models}
Different from the usual pretraining, large-scale pretraining requires distributed training techniques, including model parallelism, data parallelism, memory optimization, data synchronization, etc. Fortunately, Megatron-LM\cite{shoeybi2020megatronlm} provides us with an efficient training framework for language models. We have implemented and trained our protein language model within this framework, as well as downstream classification and regression tasks.

\section{Methodology}

\subsection{Pretrain Tasks}
\textbf{\indent Description}
\ 
\newline
The goal for protein pretraining is modeling data patterns in massive unlabeled sequences. One closed-related model is BERT\cite{bert} from natural language processing. We made some modifications to its loss function and model structure.

\textbf{Dataset}
\ 
\newline
Our work takes the dataset put forward by TAPE\cite{rao2019evaluating}. So some date descriptions are inherited. PFAM\cite{pfam2019} is a widely-used database consisting of more than 32 million protein sequences. Sequences in PFAM are clustered into evolutionarily related groups (protein families). Leveraging this family information, TAPE constructed a test set (about 1\% of the data) of fully held out families. The remaining data are used for constructing training and test sets using a random 95\%/5\% split. We use preprocessed PFAM from TAPE as the pretrain corpus.

\textbf{Training Objective}
\ 
\newline
BERT\cite{bert} original loss consists of masked language model loss and next sentence prediction loss.
% $$
% \mathcal{L}_{\text { BERT }}=\sum\mathcal{L}_{masked\ x_i}{\mathrm{MLM}}\left(x_i\right)+\sum\mathcal{L}_{sentence pair\ p_i}{\mathrm{NSP}}\left(p_i\right)
% $$

$$
\mathcal{L}_{\text { BERT }} = Loss_{MLM} + Loss_{NSP}
$$
$$
Loss_{MLM} = \sum\mathcal{L}_{masked\ x_i}{\mathrm{MLM}}\left(x_i\right)
$$
$$
Loss_{NSP} = \sum\mathcal{L}_{sentence pair\ p_i}{\mathrm{NSP}}\left(p_i\right)
$$
For protein pretraining, we inherited the masking strategy of the masked language model (MLM) in BERT\cite{bert}, randomly masking 15\% of all the amino acid tokens, and then train the protein model to be able to predict the masked token from the rest of tokens. As for next sentence prediction (NSP), considering that the input sequences are randomly shuffled, we assume there is no evident semantic/biological correlation between sequences. So we discard the next sentence prediction loss, only keep the masked language model loss.

As for the training objective function, we modified the loss function of BERT: BERT’s loss function includes masked language model and next sentence prediction. Considering that there is no obvious contextual semantic relationship between protein and protein, we only retain masked language model loss.
$$
\mathcal{L}_{\mathrm{MLM}}=-\sum_{\hat{x} \in m(\mathbf{x})} \log p\left(\hat{x} \mid \mathbf{X}_{\backslash m(\mathbf{x})}\right)
$$
In terms of model structure, Megatron-LM \cite{shoeybi2020megatronlm} proposes that when the scale of the model grows huge, the position of the layernorm becomes critical. Therefore, the sublayers in the transformer layer have been reordered. The original layernorm is in the output layer, but now it is placed ahead of the input layer to prevent the input data from drifting.

\subsection{Downstream Classification Tasks}
There are three classification tasks, corresponding to token, sequence, and token-pair classification.
\subsubsection{Secondary Structure}
\ 
\begin{figure}[ht]
  \centering
  \includegraphics[width=0.7\linewidth]{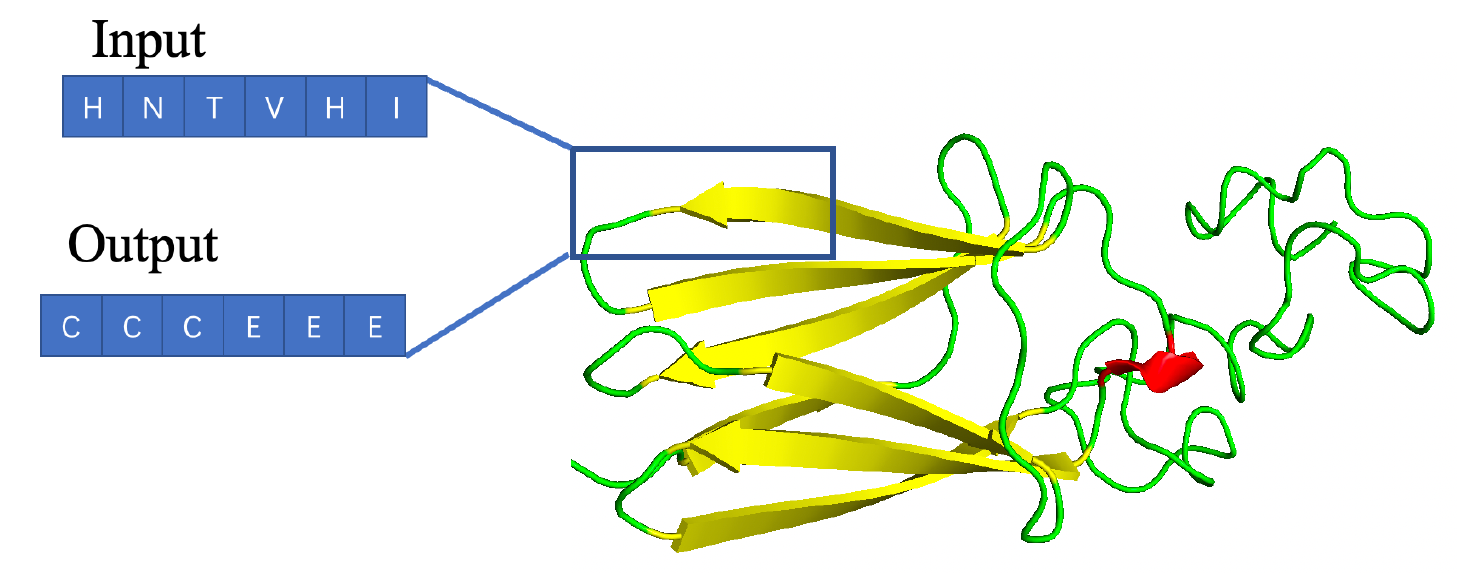}
  \caption{Secondary Structure Task}
\end{figure}

\textbf{Description}

Secondary structure classification is a token-level task. The input is protein sequence, the output is a sequence of labels with the same length, indicating the secondary structure position of the corresponding amino acid. As for Q3, the labels are Helix, Strand, and Other. As for Q8, the labels are helix (G), $\alpha$-helix (H), $\pi$-helix (I), $\beta$-stand (E), bridge (B), turn (T), bend (S), and others (C).

\textit{Input}
$$In_{seq} = (x_1, \ldots, x_L)$$

\textit{Output}
$$Out_{seq} = (y_1, \ldots, y_L). y_i \in Q_i$$

\textbf{Dataset}

The dataset for secondary structure task is the CB513\cite{cb513} dataset.

\textbf{Training Objective}

A one-dimensional convolution layer can be applied to secondary structure prediction\cite{jinbo_conv}. However, due to the powerful modeling capabilities of our model, the encoding output from the protein language model already contains sufficient information for this task, so we take ProteinLM followed by a multilayer perceptron as the secondary structure classifier.

$$
Loss(Seq) = \sum_{x_i \in Seq} CrossEntropyLoss(model(x_i), x_i)
$$

\subsubsection{Remote Homology}
\ 
\begin{figure}[ht]
  \centering
  \includegraphics[width=0.7\linewidth]{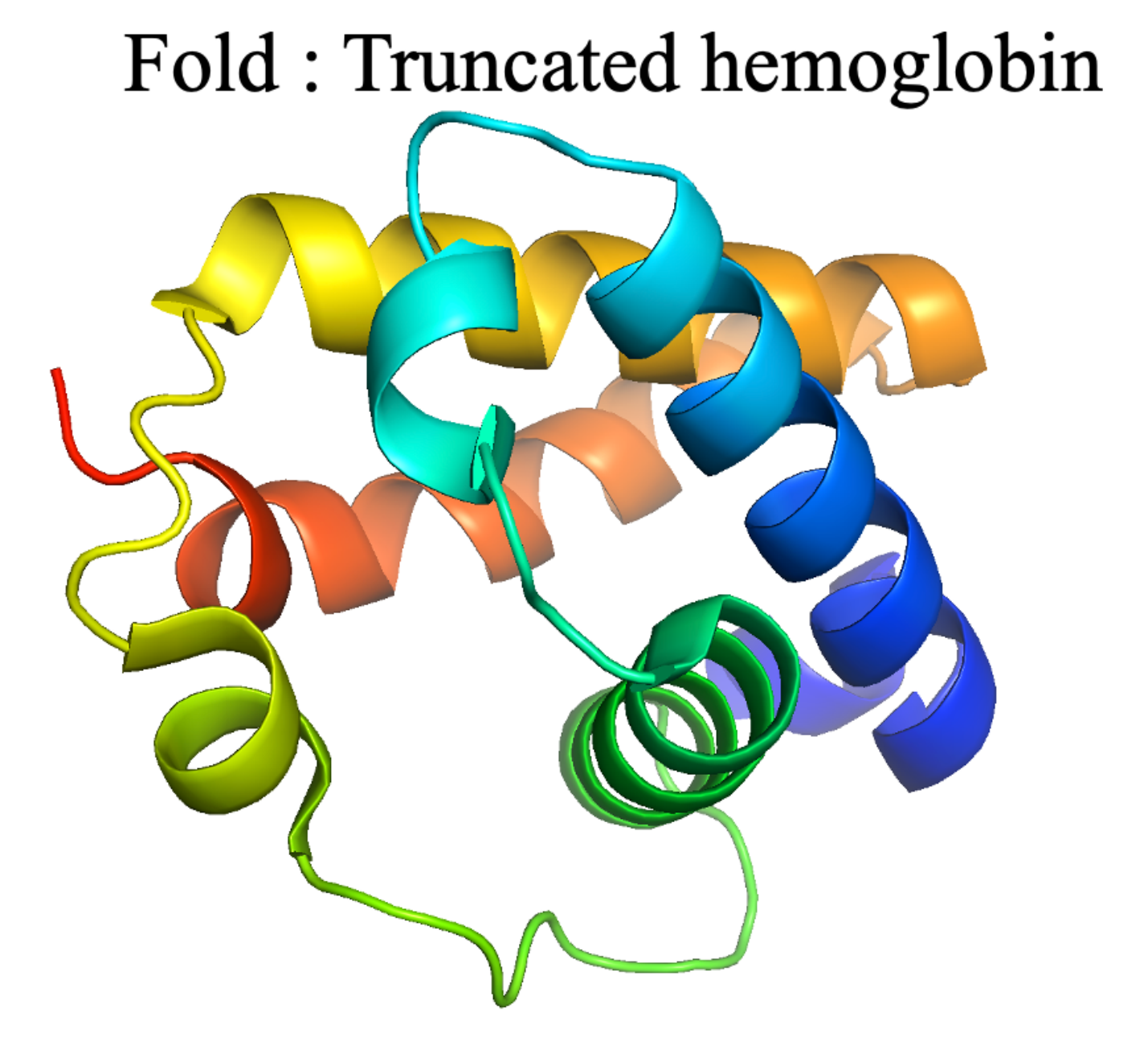}
  \caption{Remote Homology Task}
\end{figure}

\textbf{Description}

Remote homology detection is a sequence-level classification task. This task is introduced to measures a model’s ability to detect structural similarity across distantly related inputs. The input is a protein sequence, and the target is to predict which fold family this sequence belongs to. There are 1195 classes in all. Similar to the token-level prediction tasks, we adopt the multilayer perceptron for classification.
$$
Encoded\_Seq = ProteinLM(Seq(AC[1], \ldots AC[n))
$$
$$
    En([CLS], AC[1], \ldots AC[n]) = Encoded\_Seq
$$
$$
    Label(Seq(AC[1], ... AC[n)) = MLP(En([CLS]))
$$

% \begin{equation}
% \resizebox{1\hsize}{!}{$A+B+C+D+E+F+G+H+I+J+K+L+M+N+O+P+Q+R+S+T+U+V+W+X+Y+Z$}
% \end{equation}

Here, $En(x)$ means encoding results for token $x$, AC[i] means the $i^{th}$ amino acid in protein sequence.

\textbf{Training Objective}

This is a classical classification task, we take the naive cross-entropy loss.
$$
Loss(Seq) = CrossEntropyLoss(model(seq), label)
$$

\subsubsection{Contact Prediction}
\ 
\begin{figure}[ht]
  \centering
  \includegraphics[width=0.7\linewidth]{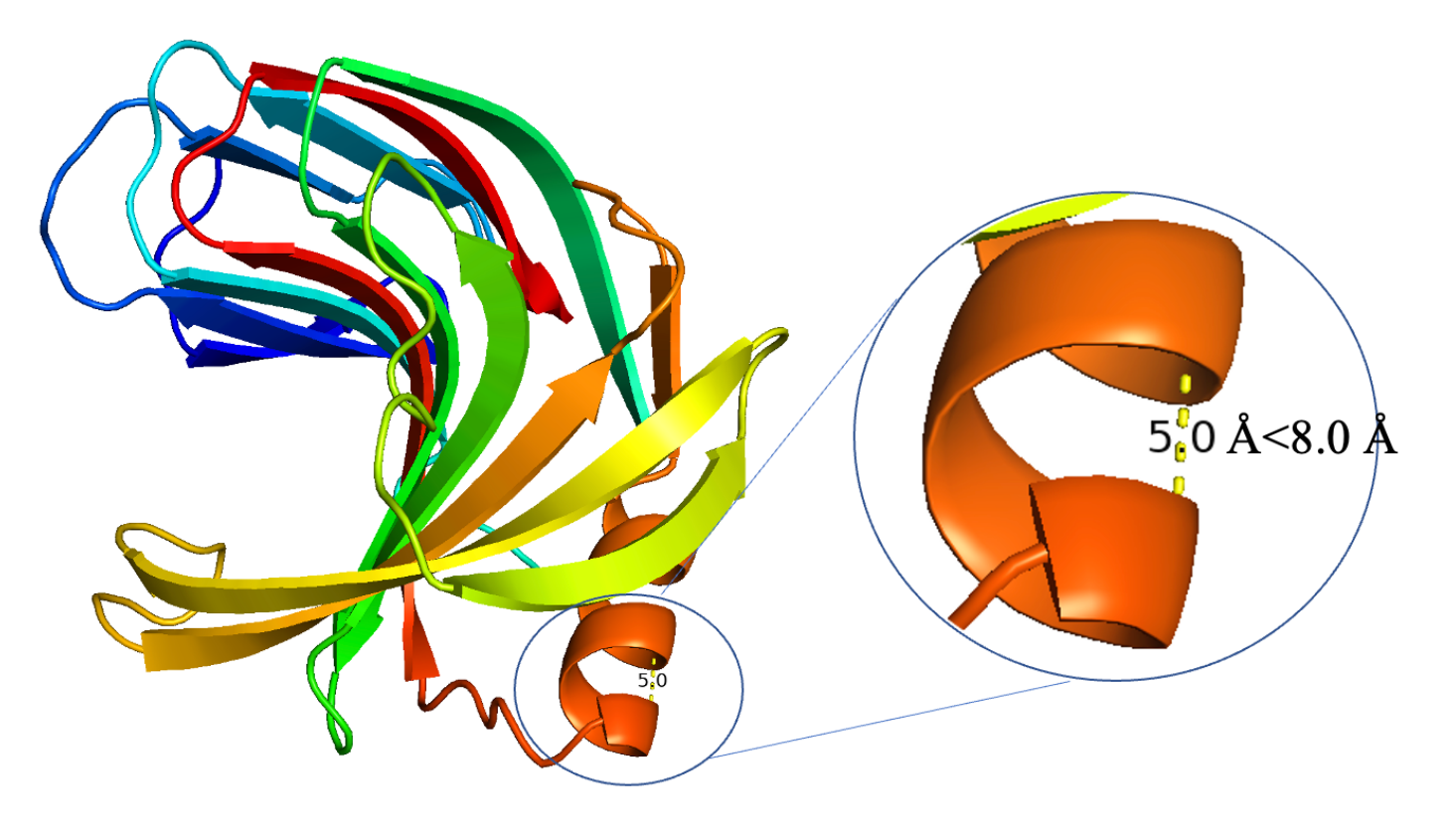}
  \caption{Contact Prediction Task}
\end{figure}

\textbf{Description}

Contact prediction means predicting whether or not amino acid pairs are in "contact" in folded structure (in "contact" means the distance in folded structure within 8 angstroms); facilitating 3-dimensional free modeling of protein. It is a classification task, assigning a binary label to amino acid pairs, indicating whether they are in 'contact'. The contact prediction task can evaluate the model's ability to capture protein sequence's global information. Unlike the commonly used residual connected 2D-convolution network, we adopted a simple predictor, concatenating embedding pairs and using multilayer perceptron to do this binary classification. Numerous hidden units, presentation layers, as well as huge corpus guarantee that our model can capture even more long-range dependence information than common models.

\textbf{Dataset}

The dataset from ProteinNet\cite{proteinnet}. And evaluation metric is $L/5$, $L/2$, $L$ most likely contact prediction accuracy contacts ($L$ is the length of protein sequence).

\subsection{Downstream Regression Tasks}

\subsubsection{Fluorescence}
\ 
\begin{figure}[ht]
  \centering
  \includegraphics[width=0.7\linewidth]{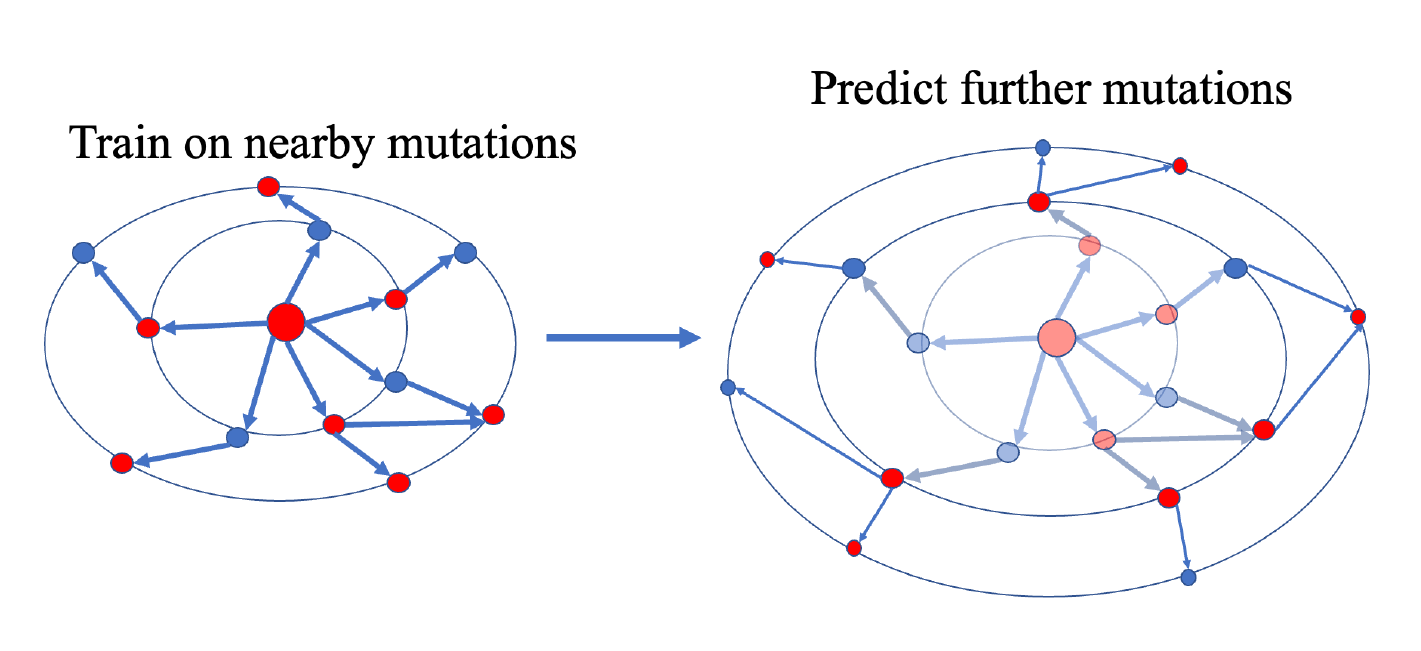}
  \caption{Fluorescence Task}
\end{figure}

\textbf{Description}

Distinguishing protein sequences with different mutations can be difficult, since the computational cost grows exponentially with the number of mutations $m$. The computational complexity for a sequence with $m$ mutation away is $ O(L^m)$. The fluorescence prediction task can evaluate the model's capacity to distinguish between very similar protein sequences, as well as its ability to generalize to unseen combinations of mutations\cite{rao2019evaluating}. Accurate predictions can facilitate the exploration of the protein landscape.

\textbf{Dataset}

The train set\cite{fluorescence} is made up of $Distance_{Hamming} = 3$ neighborhoods from the parent green fluorescent protein\cite{fluorescence}, while the test set sequences with four or more mutations.

\subsubsection{Stability}
\ 
\begin{figure}[ht]
  \centering
  \includegraphics[width=0.7\linewidth]{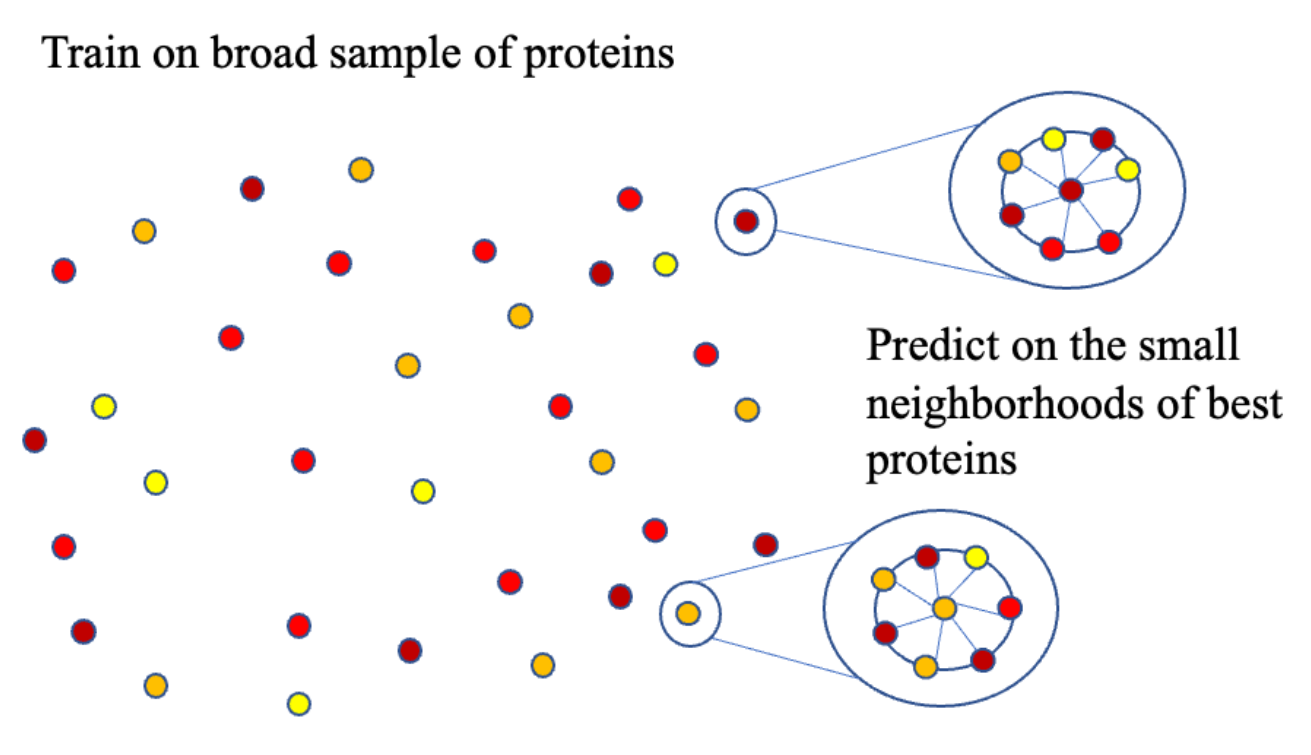}
  \caption{Stability Task}
\end{figure}

\textbf{Description}

Stability is very important in the design of protein drugs, because drugs with low stability are often degraded before they take effect. The stability of one protein sequence is measured experimentally and indirectly: the upper limit of concentration at which the protein can maintain its original folding structure\cite{stability}. Therefore, for this task, The input is the amino acid sequence, while the output is a continuous value predicting to which extent the protein can maintain its fold structure.

\textbf{Dataset}

The train set consists of proteins from four rounds of experimental design, while the test set contains Hamming distance-1 neighbors of top candidate proteins\cite{rao2019evaluating}.

\section{Results}
Our model has obtained amazing results in downstream tasks. There are great improvements in four tasks: secondary structure prediction, distant homology detection, stability, and contact prediction. It is worth mentioning that the performance of the 3B model on contact prediction has almost doubled compared with the baseline model.

Besides, we used 10 sets of model hyper-parameters in total, and conducted very sufficient experiments on a series of tasks. The results can be found in\autoref{tab:total}.
% The pretraining details can be found in \nameref{sec:supp}, and all t

\subsection{Training}

We pretrained two large models on a 480 GPUs (Tesla-V100-32GB) cluster for about three weeks. The MLM loss and PPL of the pretrained models can be found in \autoref{tab:mlm_ppl}.

The 3B parameters model reached language model loss $MLM loss_{3B} = 1.318$, perplexity $PPL_{3B} = 3.736$.

The 1.2B parameters model reached language model loss $MLM loss_{1.2B} = 1.335$, perplexity $PPL_{1.2B} = 3.802$.

In pretraining, although the overall training iteration for the 3 billion model is only half of that for the 1.2 billion model, it reached even smaller MLM loss and PPL. This phenomenon demonstrated that, when handled properly, the expansion in model scale can contribute to the accurate capture of patterns in the data.

% The training MLM loss, validation MLM loss and PPL curves can be found below.

% \begin{enumerate}
%     \item 1.2B model's training MLM loss: \autoref{fig:1.2b_loss}
%     \item 3B model's training MLM loss: \autoref{fig:3b_loss}
%     \item 1.2B model's validation MLM loss and PPL \autoref{fig:1.2b_valid}
%     \item 3B model's validation MLM loss and PPL \autoref{fig:3b_valid}
% \end{enumerate}

\begin{table}[htbp]
    \centering
    \begin{tabular}{|l|l|l|}
    \hline
        Model & Protein LM (1.2B) & Protein LM (3B) \\ \hline
        MLM Loss & 1.335 &  1.318 \\ \hline
        PPL & 3.802 & 3.736 \\ \hline
    \end{tabular}
    \caption{MLM loss and PPL}
    \label{tab:mlm_ppl}
\end{table}

% \begin{figure}
%   \centering
%   \includegraphics[width=0.8\linewidth]{pics/2048train.png}
%   \caption{1.2 Billion Model Training Loss}
%   \label{fig:1.2b_loss}
% \end{figure}

% \begin{figure}
%   \centering
%   \includegraphics[width=0.8\linewidth]{pics/3072train.png}
%   \caption{3 Billion Model Training Loss}
%   \label{fig:3b_loss}
% \end{figure}

% \begin{figure}
%   \centering
%   \includegraphics[width=0.8\linewidth]{pics/2048valid.png}
%   \caption{1.2 Billion Model Validation}
%   \label{fig:1.2b_valid}
% \end{figure}

% \begin{figure}
%   \centering
%   \includegraphics[width=0.8\linewidth]{pics/3072valid.png}
%   \caption{3 Billion Model Validation}
%   \label{fig:3b_valid}
% \end{figure}

\subsection{Evaluation}
Out of the five tasks, the results of four tasks have been improved.
\begin{enumerate}
    \item Contact Prediction: \autoref{tab:cc}
    \item Remote Homology: \autoref{tab:rh}
    \item Secondary Structure: \autoref{tab:ss}
    \item Fluorescence: \autoref{tab:fl}
    \item Stability: \autoref{tab:st}
\end{enumerate}

\begin{table}[htbp]
    \centering
    \begin{tabular}{|l|l|}
    \hline
        Task & contact prediction \\ \hline
        Metric & P@L/5 \\ \hline
        TAPE & 0.36 \\ \hline
        ProteinLM (200M) & 0.52 \\ \hline
        ProteinLM (3B) & \textbf{0.75} \\ \hline
    \end{tabular}
    \caption{Contact Prediction}
    \label{tab:cc}
\end{table}

\begin{table}[htbp]
    \centering
    \begin{tabular}{|l|l|}
    \hline
        Task & remote homology \\ \hline
        Metric & Top 1 Accuracy \\ \hline
        TAPE & 0.21 \\ \hline
        ProteinLM (200M) & 0.26 \\ \hline
        ProteinLM (3B) & \textbf{0.30} \\ \hline
    \end{tabular}
    \caption{Remote Homology}
    \label{tab:rh}
\end{table}

\begin{table}[htbp]
    \centering
    \begin{tabular}{|l|l|}
    \hline
        Task & secondary structure \\ \hline
        Metric &  Accuracy (Q-3) \\ \hline
        TAPE & 0.73 \\ \hline
        ProteinLM (200M) & 0.75 \\ \hline
        ProteinLM (3B) & \textbf{0.79} \\ \hline
    \end{tabular}
    \caption{Secondary Structure}
    \label{tab:ss}
\end{table}

\begin{table}[htbp]
    \centering
    \begin{tabular}{|l|l|}
    \hline
        Task & fluorescence \\ \hline
        Metric & Spearman's rho \\ \hline
        TAPE & 0.68 \\ \hline
        ProteinLM (200M) & 0.68 \\ \hline
        ProteinLM (3B) & 0.68 \\ \hline
    \end{tabular}
    \caption{Fluorescence}
    \label{tab:fl}
\end{table}

\begin{table}[htbp]
    \centering
    \begin{tabular}{|l|l|}
    \hline
        Task & stability \\ \hline
        Metric & Spearman's rho \\ \hline
        TAPE & 0.73 \\ \hline
        ProteinLM (200M) & 0.77 \\ \hline
        ProteinLM (3B) & \textbf{0.79} \\ \hline
    \end{tabular}
    \caption{Stability}
    \label{tab:st}
\end{table}

  \begin{table*}[hbp]
    % \begin{tabular}{lllllllll}
        \centering
        \begin{tabular}{|l|l|l|l|l|l|l|l|l|}
        \hline
            \diagbox{Model}{Performance}{Task} & P@L/5 & P@L/2 & P@L & Fluorescence & RH & SS Q@3 & SS Q@8 & Stability \\ \hline
            hidden-512-layer-32-head-8   & 0.503                   & 0.477                   & 0.409                 & 0.679          & 0.205           & 0.716                  & 0.578                  & 0.758          \\ \hline
            hidden-768-layer-12-head-6   & 0.487                   & 0.428                   & 0.369                 & 0.677          & 0.198           & 0.721                  & 0.570                  & 0.770          \\ \hline
            hidden-768-layer-16-head-16  & 0.534                   & 0.469                   & 0.396                 & 0.676          & 0.205           & 0.722                  & 0.575                  & 0.762          \\ \hline
            hidden-768-layer-16-head-24  & 0.519                   & 0.427                   & 0.376                 & 0.678          & 0.192           & 0.719                  & 0.572                  & 0.687          \\ \hline
            hidden-1024-layer-12-head-16 & 0.572                   & 0.490                   & 0.419                 & 0.676          & 0.209           & 0.729                  & 0.584                  & 0.744          \\ \hline
            hidden-1024-layer-12-head-32 & 0.500                   & 0.446                   & 0.377                 & 0.680          & 0.201           & 0.721                  & 0.575                  & 0.762          \\ \hline
            hidden-2048-layer-12-head-16 & 0.676                   & 0.576                   & 0.495                 & 0.677          & 0.266           & 0.752                  & 0.614                  & 0.732          \\ \hline
            hidden-2048-layer-24-head-16 & 0.710                   & 0.658                   & 0.563                 & 0.678          & 0.271           & \textbf{0.791}         & 0.652                  & 0.679          \\ \hline
            hidden-2048-layer-24-head-8  & 0.673                   & 0.600                   & 0.531                 & 0.674          & 0.262           & 0.762                  & 0.624                  & 0.785          \\ \hline
            hidden-3072-layer-24-head-16 & \textbf{0.753}          & \textbf{0.662}          & \textbf{0.566}        & \textbf{0.681} & \textbf{0.298}  & \textbf{0.791}         & \textbf{0.654}         & \textbf{0.794} \\ \hline
        \end{tabular}
        \caption{Results for comparative experiments.}
        $P@X$ means contact\ prediction precision@X. \\
        $SS\ Q@X$ means $X\ category$ classification for secondary structure. \\
        $RH$ means remote homology.
        \label{tab:total}
    \end{table*}

\section{Contact Map Visualization}
Generally, the accuracy of predictions on the anti-diagonal can reflect the model's ability to capture long-range dependency. Therefore, we also visualized the ground truth contact maps, as well as contact maps predicted by our model and TAPE. 
The contact map below demonstrates that our model is good at capturing long-range dependency.

\subsection{Factual Contact Map}

\begin{figure}[ht]
  \centering
  \includegraphics[width=\linewidth]{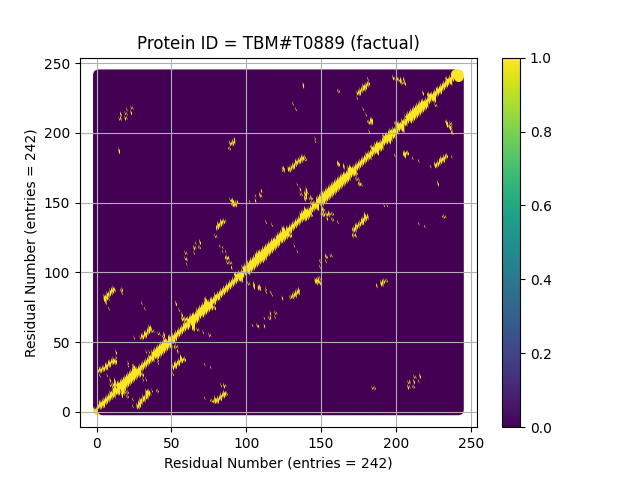}
  \caption{Factual Contact Map}
  \label{fig:fact}
\end{figure}

We visualize the contact map of protein \#TBM-T0889 in \autoref{fig:fact}, and we can intuitively see that there are many long-range contacts (contacts that are separated by at least 24 residues). This protein sequence can be used to distinguish the ability to capture long-distance dependence of different models.

\subsection{TAPE Contact Map}

\begin{figure}[ht]
  \centering
  \includegraphics[width=\linewidth]{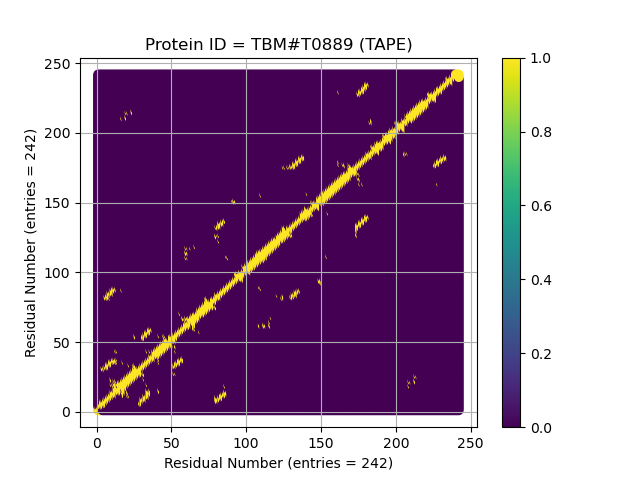}
  \caption{TAPE Prediction}
  \label{fig:tape}
\end{figure}

Through the visualized predictions from TAPE (\ref{fig:tape}), we can see that the TAPE model (small-scale transformer) can capture medium-range contacts (contacts that are separated by 12-23 residues). As for the long-range contact prediction, there are lots of missings on the anti-diagonal belt.

\subsection{ProteinLM Contact Map}
ProteinLM-3B model shows very good performance in contact prediction, and the visualized prediction map confirmed this. ProtrinLM-3B can capture medium-range and long-range dependencies, with lots of hits on the anti-diagonal belt.

\begin{figure}[ht]
  \centering
  \includegraphics[width=\linewidth]{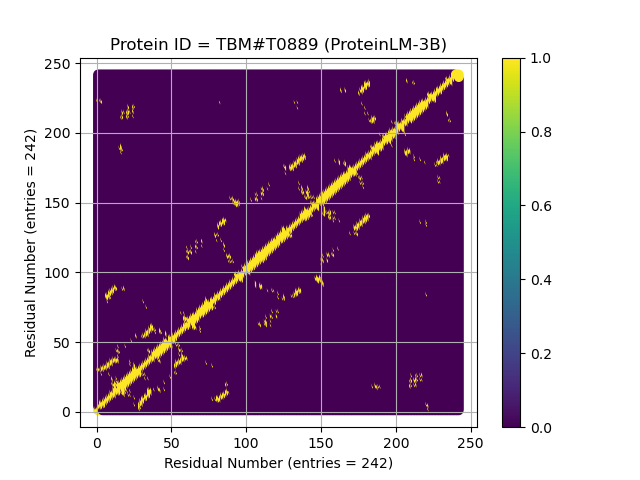}
  \caption{ProteinLM-3B Prediction}
  \label{fig:protrinlm}
\end{figure}

\subsection{Analysis and Discussion}
% In the visual contact map of our model, the contacts on the anti-diagonal line can be accurately predicted.
% The prediction results on the anti-diagonal line are more accurate.

With the limited amount of computing resources and computing time, the selection of hyper-parameters is critical to the model's performance. A trade-off is necessary. The depth (transformer layers) of the model has a greater impact on the performance than the width (hidden size). 

On the one hand, the model that is too flat ($\#transformer\_layers <8$) performs poorly, even though it has a large hidden size. We trained a model with $hidden\_size = 8192$, $\#transformer\_layers = 3$. Although its training speed (average time for each iteration) is the fastest among all models, it failed to converge after 5 days of training. 

On the other hand, the model that is too deep is not a feasible choice in this scenario (limited training time). The training time of \autoref{fig:hidden-512-layer-32-head-8-bs-16-mp-1} is 3.5 times that of \autoref{fig:hidden-2048-layer-12-head-16-bs-4-mp-1}. And it takes about 25 days to train the model with 32 transformer layers \autoref{fig:hidden-512-layer-32-head-8-bs-16-mp-1}.

Our empirical conclusion is that the model parameters of $512 < hidden\_size < 3072$, $8 < \#transformer\_layers < 24$ are feasible and can well balance training efficiency and resource consumption.

\section{Summary}
We proposed ProteinLM, a large-scale pretrain model for protein sequences. The optimizations we introduced into protein pretraining make billion-level model training possible and efficient. And the significantly improved performance in downstream tasks shows that as the scale of the model increases, the biological information in the sequence and the long-term dependence can be captured more accurately. In addition, through a large number of controlled experiments, we found and summarized some empirical rules for hyperparameter selection.

\newpage
\bibliographystyle{plain}
\bibliography{main}

\newpage
\appendix
% \section{Appendix}
\begin{figure*}[ht]
  \centering
  \includegraphics[width=\linewidth]{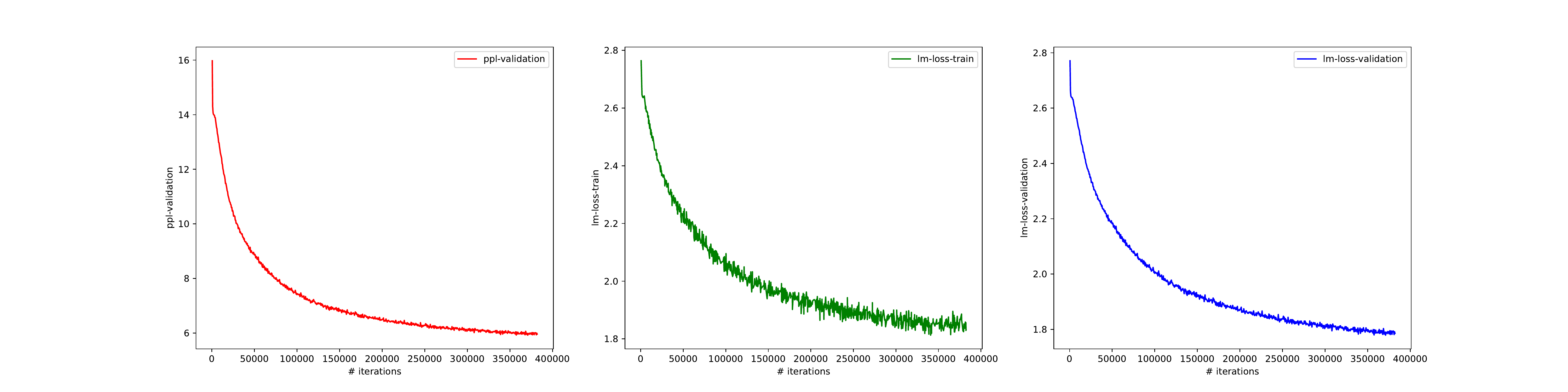}
  \caption{32 layers, hidden size = 512, 8 attention heads}
  \label{fig:hidden-512-layer-32-head-8-bs-16-mp-1}
\end{figure*}

\begin{figure*}[ht]
  \centering
  \includegraphics[width=\linewidth]{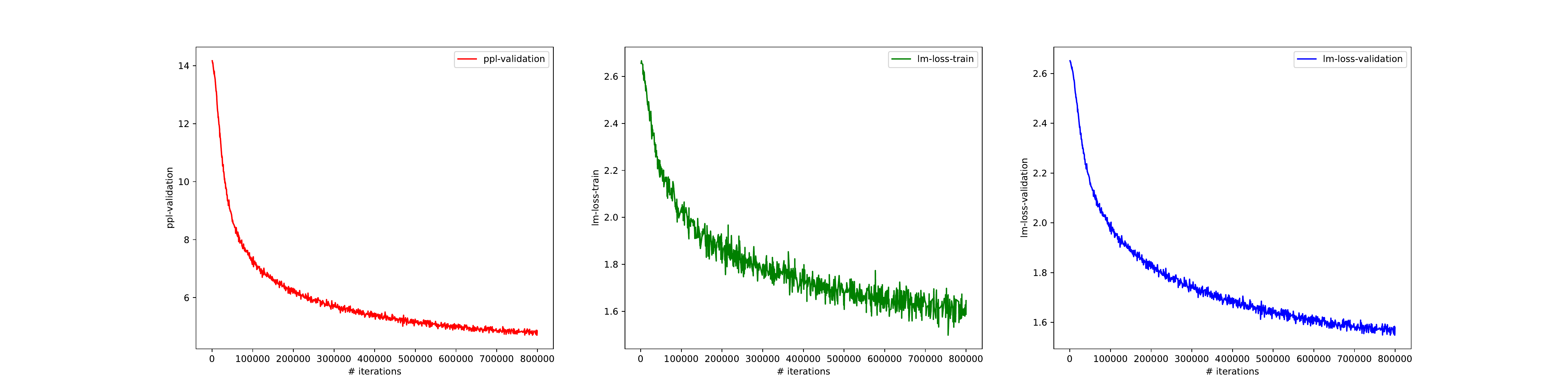}
  \caption{12 layers, hidden size = 2048, 16 attention heads}
  \label{fig:hidden-2048-layer-12-head-16-bs-4-mp-1}
\end{figure*}

\end{document}